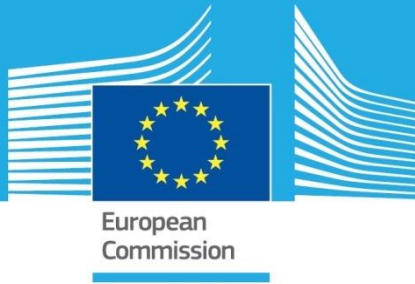

# JRC TECHNICAL REPORTS

*JRC Digital Economy Working Paper 2018-10*

# Fair and Unbiased Algorithmic Decision Making: Current State and Future Challenges

Songül Tolan

December 2018

# Fair and Unbiased Algorithmic Decision Making: Current State and Future Challenges

Songül Tolan⋆

European Commission, Joint Research Centre (JRC), Seville, Spain

**Abstract** Machine learning algorithms are now frequently used in sensitive contexts that substantially affect the course of human lives, such as credit lending or criminal justice. This is driven by the idea that 'objective' machines base their decisions solely on facts and remain unaffected by human cognitive biases, discriminatory tendencies or emotions. Yet, there is overwhelming evidence showing that algorithms can inherit or even perpetuate human biases in their decision making when they are based on data that contains biased human decisions. This has led to a call for fairness-aware machine learning. However, fairness is a complex concept which is also reflected in the attempts to formalize fairness for algorithmic decision making. Statistical formalizations of fairness lead to a long list of criteria that are each flawed (or harmful even) in different contexts. Moreover, inherent tradeoffs in these criteria make it impossible to unify them in one general framework. Thus, fairness constraints in algorithms have to be specific to the domains to which the algorithms are applied. In the future, research in algorithmic decision making systems should be aware of data and developer biases and add a focus on transparency to facilitate regular fairness audits.

⋆ songul.tolan@ec.europa.eu - I thank Emilia Gómez, Marius Miron, Carlos Castillo, Bertin Martens, Frank Neher and Stephane Chaudron for their thoughtful comments. This article is part of the HUMAINT research project: `https://ec.europa.eu/jrc/communities/community/humaint`

# Table of Contents

# 1 Introduction

## 1.1 Context

We review the problem of discrimination and bias in algorithmic decision making. Data-driven algorithmic decision making (ADM) systems are now frequently used in sensitive contexts that substantially affect the course of many human lives, such as credit lending, health or criminal justice. This is driven by the idea that 'objective' machines base their decisions solely on facts and remain unaffected by human cognitive biases, discriminatory tendencies or emotions. Yet, there is overwhelming evidence showing that algorithms can inherit or even perpetuate human biases in their decision making when their underlying data contains biased human decisions (Barocas and Selbst, 2016). This has led to a call for fairness-aware machine learning. However, fairness as a complex value-driven concept is hard to formalize for the use in ADM systems. Statistical formalizations of fairness lead to a long list of criteria that can be useful in one context but can be flawed (or harmful even) in different contexts or in the presence of bias. Moreover, tradeoffs in these criteria make it impossible to unify them in one general framework.

We cover the state of the art in evaluating discrimination and bias in algorithmic decision making. That is, we present common fairness evaluation methods in ADM systems, discuss their strengths and weaknesses and discuss the tradeoffs that make it impossible to unify these methods in one general framework. Moreover we discuss how unfairness appears in algorithms in the first place. We conclude with the notion that there is no simple one-size-fits-all solution to fairness in machine learning. Researchers who aim at developing fair algorithms will have to take into account the social and institutional context, as well as the consequences that arise from the implementation of the technology. Finally, we discuss promising research avenues in addressing this topic in the future and identify practices that help ensuring fairness in algorithmic decision making.

We focus on ADM systems on the basis of machine learning (ML). Fairness is a relevant issue in other ML applications such as recommender systems or personalized ad targeting but we focus on the case of supervised ML classification algorithms for ADM. ML classification algorithms have the potential to support in decision making situations which may have an impact on other individuals or the society as a whole. In many cases, these decisions are linked to a prediction problem at their core. For instance, a school must decide which teacher to hire among a number of applicants. "Good" teachers have a positive impact on student performance but student performance cannot be observed when the hiring decision has to be made. Thus, the decision maker has to "predict", based on observed characteristics, which applicant is likely to belong to the category of "good" teachers. Over the last decades, the availability of data related to such decision-making processes has provided quantitative ways to exploit these data sources and inform human decision-making, e.g. by means of (ML) algorithms that can learn from this data and make predictions based on it. We illustrate the idea ML classification in Figure 1. The rules of ML classifica-tion algorithms are learned from datasets on categorized (i.e. labelled) observations to identify the membership of new observations to given categories. For instance, a classification algorithm can use a dataset of images labeled as dog and cat pictures to identify whether a new image belongs to the dog or cat category. Similar to this, ADM systems use present data to inform decision makers by predicting the probability (or risk) that a present decision object belongs to a outcome-category that is relevant to the decision context.

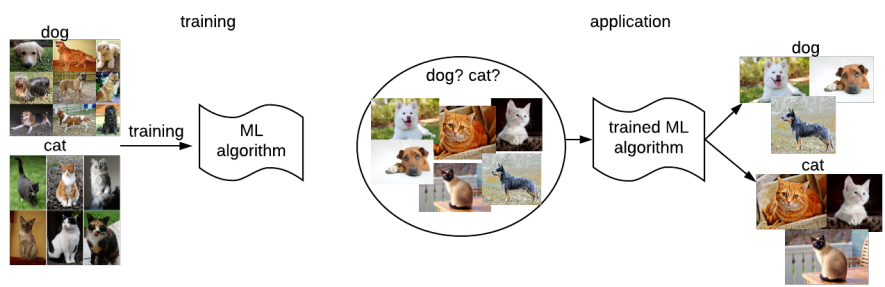

Fig. 1: Illustration of a classification algorithm

Note that we distinguish between bias and (un-)fairness. While we try to identify and mitigate both, there is also a fundamental difference: fairness is a normative concept while bias is a technical concept. The literature on fair ML algorithms mainly derives its fairness concepts from a legal context. Generally, a process or decision is considered fair if it does not discriminate against people on the basis of their membership to a protected group, such as sex or race.[1] The right to non-discrimination is embedded in the normative framework of the EU. Explicit mentions of it can be found in Article 21 of the EU Charter of Fundamental Rights, Article 14 of the European Convention on Human Rights, and in Articles 18-25 of the Treaty on the Functioning of the European Union (Goodman and Flaxman, 2016). In this context we distinguish between common legal definitions of fairness, direct discrimination (or *'disparate treatment'*) and indirect discrimination (or *'disparate impact'*) as well as definitions that can be derived from the legal framework but are based on ADM system scores. Direct discrimination occurs when a person is treated differently based on her membership to a protected group. Indirect discrimination occurs when an apparently neutral rule leads to outcomes that differ based on a persons membership to a protected group (European Union Agency for Fundamental Rights; European Court of Human Rights;Council of Europe;, 2018). Discrimination based on scores occurs when the risk scores of an ADM system are not equally predictive of the actual outcomes for different protected groups.

---

[1] For instance, Article 14 of the European Convention on Human Rights states as protected group "sex, race, colour, language, religion, political or other opinion, national or social origin, association with a national minority, property, birth or other status".

In contrast, bias is a systematic deviation from a true state. From a statistical perspective an estimator is biased when there is a systematic error that causes it to not converge to the true value that it is trying to estimate. In humans, bias can manifest itself in deviating perception, thinking, remembering or judgment which can lead to decisions and outcomes differing for people based on their membership to a protected group. There are different forms of bias, such as the subjective bias of individuals, data bias, developer bias and institutionalized biases that are ingrained in the underlying societal context of the decision. Bias, if not controlled for, can cause unfairness in ADM systems.

In what follows, we first motivate our topic. We start our analysis with a presentation and discussion of definitions of algorithmic fairness. In Section 3 we discuss sources of discrimination. In Section 4 we elaborate on the future work necessary to overcome challenges towards fairness-aware ADM systems. Section 5 concludes.

### 1.2 Fairness and bias in humans and algorithms

Since 2012 the study of fairness in machine learning has been growing rapidly. The topic also remains in the public media due to frequent articles in major news and science outlets, such as Financial Times[2] or Nature[3]. In 2018's leading international machine learning conference (ICML) the topic on fairness in machine learning covered several conference sessions, tutorials and critical podium discussions. By now, there are many computer science conferences dedicated to the topic of fairness in algorithms. This not only shows that this is a 'hot topic', it also emphasizes the relevance of this research domain as more and more algorithmic decision systems are being deployed frequently in highly sensitive areas that affect the course of human lives and society as a whole. For instance, in credit lending algorithms are being used to predict the risk of credit applicants defaulting (Huang et al., 2007). Employers use ADM systems to select best applicants. Machine learning can also be used to predict mortality risk of acute patients to improve the targeting of palliative care (Avati et al., 2017). Or, in criminal justice algorithms are being deployed to inform judges about the flight risk and re-offense risk of defendants (Kleinberg et al., 2017; Angwin et al., 2016).

The contexts in which ADM systems are being deployed have one thing in common: the decisions have prediction problems at their core. Machine learning is a discipline that aims at maximizing prediction performance (Kleinberg et al., 2015). The idea of an 'objective' machine making decisions solely based on facts in areas where decisions affect many people's lives might be appealing. Especially if we are aware of the subjective biases that human decision makers are prone to (Kahneman, 2011). Human biases in sensitive areas of decision making are a serious problem. Multiple studies show how extraneous factors, such as mood or hunger, can significantly affect judicial decision making (Danziger et al., 2011; Chen et al., 2016). Compared to this, with enough data that is representative of the respective decision problem machine-learning algorithms can make predictions that are as good as or even more accurate than human expert predictions. However, due to the opacity of many ADM systems (because of black-box approaches or secrecy around proprietary algorithms) (Miron, 2018), it is often difficult to have a transparent description of how a machine decision was made. As a consequence, it is difficult to ensure that these algorithms are free of biases and that they adhere to the respective standards of fairness. There is overwhelming evidence showing that algorithms can inherit or even perpetuate human biases in their decision making when they are trained on data that contain biased human decisions (Barocas and Selbst, 2016).

---

[2] https://www.ft.com/content/46def3ba-8f67-11e8-9609-3d3b945e78cf
[3] https://www.nature.com/articles/d41586-018-05469-3

Indeed, ADM systems can discriminate, as seen in an article titled "Machine Bias" (Angwin et al., 2016) by the investigative news organization ProPublica[4]. The authors analyzed the re-offense risk assessment tool COMPAS which is being deployed in federal US criminal justice systems. Angwin et al. (2016) stated that this tool was "biased against blacks" as among the defendants who did not re-offend in the two year window of analysis blacks were more than twice as likely to be classified by COMPAS as medium or high risk of re-offense as whites (42 percent vs. 22 percent). Northpointe (the company that created COMPAS) rejected ProPublica's reproach of discrimination by arguing that their score was "calibrated" as blacks and whites with equal risk scores have practically the same probability of actual re-offense. Subsequent research shows that it is mathematically impossible to satisfy both notions of fairness if the true recidivism probability differs between both groups. In the following, we elaborate on this dispute and discuss the research that followed it.

## 2    Algorithmic fairness

Principally, in a legal context there is fairness when people are not discriminated against based on their membership to a (protected) group or class. In practice there are several definitions of algorithmic fairness that try to achieve this goal. In fact, in the literature we talk of at least 21 definitions of fairness (Narayanan, 2018). Many of these definitions are also surveyed and discussed in Berk et al. (2017) or Narayanan (2018). We distinguish between two categories of fairness: individual fairness and group fairness (also known as statistical fairness). In addition, we distinguish between three categories of statistical fairness definitions that are based on: predicted classifications, a combination or predicted classifications and actual outcomes, and a combination of predicted risk scores and actual outcomes. As the more applicable definition in algorithmic decision making, group fairness has been studied more extensively in the literature. For illustrative purposes we present these definitions in the classification context of re-offense prediction in criminal justice with a binary outcome (high and low re-offense risk) but most of the following definitions can also be extended to a context with multiple outcomes. Thus, we present the case of a ML algorithm that predicts the risk $R$ that a defendant will re-offend. We denote the outcome re-offense as $Y$, where $Y = 1$ if the defendant re-offended. The predicted outcome is represented by $\hat{Y}$. The ML algorithm classifies someone as high risk for recidivism, i.e. $\hat{Y} = 1$ if the risk score $R$ surpasses a predefined threshold ($r_t$), i.e. $R > r_t$. We further observe information on characteristics (or features) of a defendant in a dataset. We denote the matrix of these features $X$. We further observe the protected attribute $A$, such as gender or race.

---

[4] https://www.propublica.org/article/machine-bias-risk-assessments-in-criminal-sentencing

## 2.1 Statistical/group fairness

Statistical fairness definitions are best described with the help of a confusion matrix which depicts a cross-tabulation of the actual outcomes ($Y$) against the predicted outcomes ($\hat{Y}$), as shown in Table 1. The central cells report counts of correct and wrong classifications of the algorithm.

Table 1: Confusion Matrix

|  |  | Predicted Classification |  |  |
|---|---|---|---|---|
|  |  | $\hat{Y} = 1$ | $\hat{Y} = 0$ |  |
| Outcome | $Y = 1$ | True Positives (TP) | False Negatives (FN) | False Negative Rate (FNR) $FN/(TP+FN)$ |
|  | $Y = 0$ | False Positives (FP) | True Negatives (TN) | False Positive Rate (FPR) $FP/(FP+TN)$ |
|  |  | False Omission Rate (FOR) $FP/(TP+FP)$ | False Discovery Rate (FDR) $FN/(FN+TN)$ |  |

Cross-tabulation of actual and predicted outcomes.

The correct (*True*) classifications are reported in the diagonal of the central table and the incorrect (*False*) classifications in the off-diagonal of the central table. Different error rates are presented in the last row and the last column. False negative rate (FNR) and false positive rate (FPR) are defined as fractions over the distribution of the true outcome. False discovery rate (FDR) and false omission rate (FOR) are defined as fractions over the distribution of the predicted classification. Using this setup we now describe three broad categories into which the different definitions of group fairness fall.

### 2.1.1  Based on predicted classifications

This category of group fairness is also known as **demographic parity** and is based on predicted classifications ($\hat{Y}$, see also the columns in Table 1). Demographic parity is fulfilled if the following condition holds:

$$\mathbb{E}[\hat{Y} = 1 \mid A = a] = \mathbb{E}[\hat{Y} = 1 \mid A = b] \tag{1}$$

meaning that the share of defendants classified as high risk should be equal across different protected groups. In other words, people from different protected groups should have on average equal classifications. This definition can also be extended to *conditional* **demographic parity**:

$$\mathbb{E}[\hat{Y} = 1 \mid X = x, A = a] = \mathbb{E}[\hat{Y} = 1 \mid X = x, A = b] \tag{2}$$

which states that the share of defendants classified as high risk should be equal for defendants with the same realizations in their 'legitimate' characteristics $X$ but different realizations for the protected attribute $A$. Where 'legitimate' features are characteristics that are considered OK to discriminate against, as they are highly predictive of the outcome but not of the protected attribute. Put simply, people with the same legitimate characteristics from different protected groups should on average have the same classification. This is also related to affirmative action which is particularly relevant in recruitment decisions.

The problem with this fairness definition is the missing link to the actual outcome, which could cause the problem of the *self-fulfilling prophecy* or *reverse tokenism* Dwork et al. (2012). In case of the self-fulfilling prophecy the decision maker could fulfill demographic parity by classifying the same share of people across protected groups as low risk but she could select truly low risk people from one group and make a random selection from the other group causing a bad track record for the group with random selection. This situation could occur intentionally but it can also happen if prediction performance of the classifier is better for one group than the other. For instance, if women were released more often than men, we would have more observations for women than for men on which we could train a classifier with a better predictive performance, causing even worse outcomes for released men as we continue to apply this classification algorithm. In case of reverse tokenism, the decision maker could intentionally not release low risk females as 'tokens' to also not release other males with equal risks. Finally, demographic parity implicitly assumes that there are no intrinsic differences between different protected group features, i.e. between men and women or people of different races. This should hold despite the fact that people from different protected groups are often exposed to very diverging history that had in impact on the societal composition of each respective group. Naturally, one might conclude that this is a very strict assumption which often does not hold in reality.

### 2.1.2 Based on predicted and actual outcomes

This definition of fairness relates to the principle of indirect discrimination. In this case we look at both predicted and actual outcomes which can be deducted from the individual cells in Table 1. ProPublica's approach to discover discrimination in COMPAS falls within this category. Following ProPublica's argumentation this definition can be expressed in other words as follows. If the classifier gets it wrong, it should be equally wrong for all protected groups, since being more wrong for one group would result in harmful (or beneficial) outcomes for this group compared to the other group. This would be discriminatory. An equivalent statement can be made for correct classifications. Formal fairness definitions of this category have been made by Hardt et al. (2016); Chouldechova (2017) and Zafar et al. (2017a). For instance **error rate balance** as in Chouldechova (2017) is given if the following conditions hold.

$$\mathbb{E}[\hat{Y} = 1 \mid Y = 0, A = a] = \mathbb{E}[\hat{Y} = 1 \mid Y = 0, A = b] \qquad (3)$$

and

$$\mathbb{E}[\hat{Y} = 0 \mid Y = 1, A = a] = \mathbb{E}[\hat{Y} = 0 \mid Y = 1, A = b] \qquad (4)$$

Meaning that FNR and FPR (see Table 1) should be equal across different protected groups. Chouldechova (2017) also shows how unbalanced error rates can lead to harmful outcomes for one group in the case of re-offense prediction as predictions of higher re-offense risk lead to stricter penalties for the defendant. Equivalent formalizations can be made for other error rates as well as conditions on true rates (e.g. equal TPR, defined as "equal opportunity" in Hardt et al. (2016)).

Problematically for this category of fairness definitions, Corbett-Davies et al. (2017) show that any statistic from a confusion matrix (which is based on counting correct and wrong classifications) can be manipulated through (intentionally) harmful external changes to the real-world processes that are reflected in the data. For instance, in the case of rec-offense risk prediction, the FPR (see Table 1) can be lowered by arresting more innocent people and classifying them as low risk. The relevance of this problem, called *infra-marginality* is highlighted in the case of police stop practices in North Carolina (Simoiu et al., 2017).

### 2.1.3 Based on predicted risk scores and actual outcomes

This category of fairness definitions is commonly known as **calibration**. While the other group fairness definitions relate to the fairness of classifications, calibration[5] relates to the fairness of risk scores ($R$). Calibration means that for a given risk score $R = r$, the proportion of people re-offending is the same across protected groups. Formally, the following condition has to hold for calibration:

$$\mathbb{E}[Y = 1 \mid R = r, A = a] = \mathbb{E}[Y = 1 \mid R = r, A = b], \forall r \in R]  \quad (5)$$

---

[5] Also known as predictive parity (Chouldechova, 2017)

This is also the fairness criteria that Northpointe (the company that owns COMPAS) shows to fulfill for the COMPAS risk score. In fact, they reject ProPublica's discrimination reproaches by showing that COMPAS is calibrated for race.[6] Clearly, the two fairness criteria are in conflict, as we discuss in Section 2.4.

Barocas et al. (2018a) show that calibration is often satisfied by default if the data on which the classifier is trained on ($X$) contains enough other features that predict the protected features in $A$. This also means that calibration as a fairness condition would not require much intervention in the existing decision making processes. Moreover, calibration, too can be manipulated and there is no way to identify manipulation if we only have access to the score ($R$) and the outcome ($Y$). Corbett-Davies et al. (2017) show how a risk score can be manipulated to appear calibrated by ignoring information about the favoured group which they relate to the historical practice of redlining in credit risk assessments.

### 2.2   Individual fairness

Except in the case of conditional demographic parity, group fairness definitions only aim to ensure equality between group averages. In contrast, individual fairness takes into account additional characteristics of individual observations ($X$) and looks, as the name says, at differences between individuals rather than groups. It is driven by the notion that *similar people should be treated similarly*. Similarity depends on a function that determines the distance between two individuals ($i, j$) in terms of their predicted outcomes ($\hat{Y}_i, \hat{Y}_j$) and individual characteristics ($X_i, X_j$). Individual fairness is given if the distance between the predicted outcomes is not greater than the distance between the individual characteristics (Dwork et al., 2012). In a sense conditional parity is a more specific case of the more generic definition of individual fairness and both definitions come with the same problem as they rely on the choice of features for the vector $X$. This is not a trivial task and Dwork et al. (2012) leave this question open for further research. The non-triviality of choosing the right attributes is also reflected in the following section.

### 2.3   Direct vs. indirect discrimination

At a first glance it seems that we can fix the problem of algorithmic discrimination by restraining the algorithm from taking into account the protected group features, such as gender or race. That is, we impose on the decision making algorithm to remain blind towards sensitive attributes. In practice, this relates to the tension between direct and indirect discrimination. In EU Law both the right to direct and indirect non-discrimination are two highly held principles. The breach of either one of these principles is only justified in exceptional cases and can only be decided upon on a case-by-case basis (Fribergh and Kjaerum, 2011).

---

[6] Although, COMPAS is not calibrated for gender (Corbett-Davies et al., 2017).

The tension between these two principles is a common problem in the field of machine learning. For instance, Chouldechova (2017) shows that if the distribution of risk is different for different protected groups, then adjusting for the criteria of balanced error rates requires the setting of different classification thresholds ($r_t$) for different groups; i.e. holding people from different groups to different standards to achieve equal outcomes. Or Corbett-Davies et al. (2017) show that COMPAS is not calibrated between men and women as women have a lower re-offense prevalence than men for the same risk score. They argue that adjusting for the protected feature 'gender' would help to adjust the risk score for this type of discrimination. Generally, many computer scientist argue for the necessity of using information on sensitive attributes to adjust for fairness in data-driven decision making (Žliobaitė and Custers, 2016)

In any case, keeping a classifier blind to sensitive attributes is not a useful strategy to prevent indirect discrimination. In practice discrimination can still feed into a data driven decision system if the protected group features are correlated with other features or the predicted outcome itself. In the presence of big data, this is almost always the case (Hardt, 2014; Barocas and Selbst, 2016). In order to circumvent this problem other researchers suggest to use the protected group feature during the training of the classifier but not when it is used for prediction (Zafar et al., 2017b). But Lipton et al. (2018) show that this practice does not prevent disparate impact if other features are predictive or partly predictive of group membership. In the context of credit risk classification, Hardt et al. (2016) quantify by how much different fairness constraints are breached as race blindness is imposed on the classifier.

Thus, an overwhelming part of the computer science literature supports the view, that blindness to sensitive attributes is not a good way to account for fairness in algorithms. Yet, the in May 2018 implemented EU General Data Protection Regulation (GDPR) specifically addresses discrimination through profiling (or algorithm training) on the basis of *'sensitive data'*, where sensitive data could at least cover the sensitive group features directly, such as gender or race or at most cover features correlated with the sensitive group features, such as the postal code (Goodman and Flaxman, 2016). In the context of data-driven decisions in sensitive areas such as credit lending, it is also completely understandable that individuals may not want to trust credit providers with their sensitive data if they might as well use it with malicious intentions. It may also put their privacy at risk. Therefore, Kilbertus et al. (2018) propose a protocol from secure multi-party computation between individual users, providers of data-driven services and a regulator with fairness aims. This method allows for fairness certification and fairness adjustments of ADM systems while keeping sensitive attributes encrypted to both, the regulator and the ADM system provider. Yet, legally implementing such protocols poses additional challenges as we would have to decide on who should perform such audits and who should carry the costs of these audits (Goodman, 2016b).

## 2.4 Tradeoffs

There are several methods to ensure that a classifier fulfills the above mentioned fairness criteria. In most cases, adjusting for fairness is seen as a constrained optimization problem where the difference between the actual outcome ($Y$) and the predicted classification ($\hat{Y}$) is minimized subject to the constraints of different fairness criteria. Clearly, imposing too many restrictions to the classifier at once will make it impossible to find a meaningful solution.

In fact, several mathematical proofs show the incompatibility of these fairness criteria under fairly weak assumptions. Most prominently, Kleinberg et al. (2016) and Chouldechova (2017) mathematically proof that it is impossible to reconcile calibration and fairness criteria based on a combination of predicted and actual outcomes (Section 2.1.2) if the prevalence of the outcome differs across different protected groups. Berk et al. (2017) illustrates the seriousness of this result. It means that any one classifier can never be adjusted for all fairness criteria if prevalence is different for different groups. Instead, the choice of the fairness criteria depends on the respective moral context of the decision. That is, this finding takes away the notion of objectivity in adjusting ADM systems for fairness.

Finally, Corbett-Davies et al. (2017) show that there also exists a trade-off between these common measures of fairness and utility. In their criminal justice case, utility depends on public safety (keep as many dangerous people in prisons as possible) and the public costs of incarceration (people in prison, especially innocent people in prison impose costs on society). Compared to a classifier that does not specifically account for fairness, different classifiers adjusted to different fairness risk criteria lead to over-proportional decreases in public safety or increases in public costs.

To sum up, the problem of algorithmic discrimination has no one-size-fits-all solution as different fairness definitions have different meanings in different contexts and not all fairness criteria can be simultaneously fulfilled in one decision process. In addition, every decision process involves different stakeholders: decision-makers, individuals affected by the decision, and sometimes the general public as some decisions (such as in the case of criminal justice) also have an impact on the society as a whole. Navigating these tradeoffs means that decisions on algorithmic fairness criteria have to be made on a case by case basis. Getting from formal abstract definitions, to a meaningful fairness context means defining an answer to the following questions in each case:

- Who are the stakeholders involved?

- What is the role of each stakeholder?
- What are the immediate costs for each stakeholder?
- What are the long term costs or consequences of the decision taken for each stakeholder?
- Do legal/moral foundations already exist?

For instance, in the case of criminal justice, the cost-benefit analysis between safety, short-term costs of incarceration as well as long term consequences of incarceration (Corbett-Davies and Goel, 2018) provide a good starting point for a meaningful context in which to discuss fairness. However, this should be extended by taking into account the differences in costs for different stakeholders, such as the general public, institutions and decision makers as well as the people to whom these decisions are applied to. In addition, fairness restrictions to the decision should adhere to already existing legal and moral foundations in each context. Domain experts can help in the definition of this meaningful context.

Nevertheless, all these definitions remain meaningless if we do not take into account the different sources of discrimination in algorithms. Only when we understand the underlying mechanisms that lead to discrimination in algorithms can we ensure that adjusting for fairness will lead to overall improvements. These mechanisms will be discussed in more detail in the following section.

## 3 Sources of discrimination

One of the major obstacles towards fair machine learning is the presence of bias. Using machine learning for ADM systems in contexts that affect people is per construction prone to many sources of bias. Figure 2 illustrates where bias can occur in the process of developing a ML algorithm.

For instance, we know that humans can have their individual biased preferences (Goodman, 2016a) or that we are prone to cognitive biases (Kahneman, 2011). If the same bias is shared by many individuals, it manifests itself in institutionalized bias. In addition, machine learning is data driven and data is per construction a reductive description of the real world. Naturally, data can never capture all real world processes (Calders and Žliobaitė, 2013). We distinguish between biases that are already in the data due to biased data-generating processes, biases that occur during the development of the algorithm due to biased developers and other human biases that can be reflected in the data but that are more difficult to define in a pure fairness and machine-learning context.

### 3.1 Bias in data

Barocas and Selbst (2016) elaborate on the different ways that data (and in particular 'big data') can be biased. For instance, since data for decision making represents a collection of past decisions, it will also inherit the prejudices of prior decision makers. In addition, decision makers are also affected by existing biases in society and these will be reflected in the data, too. In other cases big data contains correlations relevant for decision making that are in fact only based on patterns of exclusion and unequal treatment.

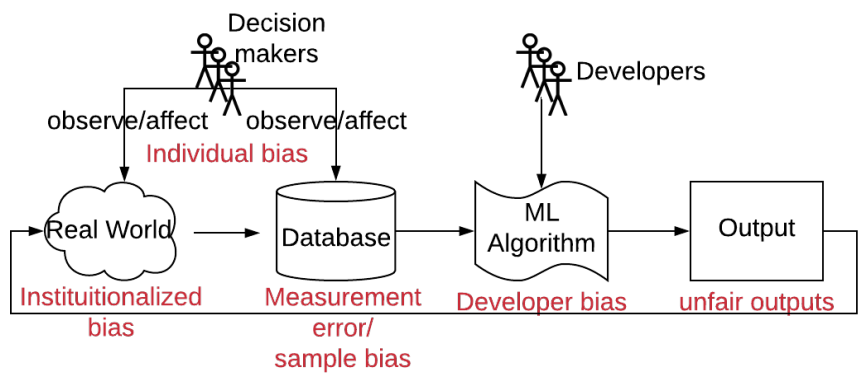

Fig. 2: Sources of unfairness in machine learning

The problem of **selective labels** (Kleinberg et al., 2017) occurs due to the nature of data used for the generation of ADM systems as records from previous human decisions. Then, we often only observe the outcome (or label) of one side of the decision. For instance, in health we only observe the health outcomes of patients who were assigned a treatment. Or in judge bail decisions we only observe crimes committed by released defendants, not of jailed defendants (Kleinberg et al., 2017). Predicting crime rates for the jailed is problematic as judges might have selected these individuals based on features not observed in the data, thus creating biased machine predicted outcomes based on observables. There would be no problem if researchers could observe what would have happened, had the jailed been released. However, this is the 'counterfactual' scenario that does not occur in the real world and thus, can never be observed. The authors address this problem by applying a technique from the econometric literature (as this literature mostly deals with causal inference) called 'contraction' (Lakkaraju et al., 2017) in which the researchers exploit the random allocation of defendants to judges. An alternative approach would be to assume that judge decisions could be mostly explained by the features observed in the data, i.e. a 'selection on observables' assumption combined with a sensitivity analysis as conducted by Jung et al. (2017) and similarly combined with a Bayesian approach by Jung et al. (2018b) and Jung et al. (2018a). These identification techniques are drawn from the econometric causal inference literature. These approaches require a deeper understanding of the data-generating process and the institutional background of the data which heavily increases the workload invested in evaluating an algorithm for fairness. However, this investment is valuable since ignoring this bias in the data would lead to biased results.

Another problem is **sample bias** which occurs when the data sample on which the algorithm is trained for is not representative of the overall population due to a systematic error in data collection. Any decision making system based on this sample would be biased in favour or against the over- or underrepresented group (Barocas and Selbst, 2016; Drosou et al., 2017; Chouldechova and Roth, 2018). There are many reasons why a group would be over- or underrepresented in the data.

One of this reasons is **measurement error**. For instance, in the criminal justice case, the risk of re-offense is measured through re-arrest. However, this could be a poor proxy for actual crimes committed if one protected group has a higher likelihood of being arrested than the other. For instance, predictive policing that directs higher police awareness to neighbourhoods with a higher share of one protected class could also lead to over-proportional arrests of this group. So far, the literature has only come up with one way to address this and other problems of measurement error: improve measurement. For instance, Corbett-Davies et al. (2017) address mismeasurement of crimes committed by only counting incidences of violent crime as evidence suggests that this constitutes a better proxy for actual re-offense. Mullainathan and Obermeyer (2017) show how the prediction of strokes (in a medical setting) can be biased due to mismeasurement of the stroke event if hospital visits are used as a proxy. This could have far-reaching societal implications as marginalised people or just those that cannot afford health insurance would be less likely to visit a hospital.

In addition, sample bias can also have deep societal roots. Especially minorities are often vulnerable to being underrepresented in relevant datasets. This point has also been made by Lerman (2013) as he discusses the problem of "the nonrandom, systemic omission of people who live on big data's margins, whether due to poverty, geography, or lifestyle, and whose lives are less 'datafied' than the general population's." Crawford (2013) also illustrates another case of under-representation in data related to the smartphone app 'Street Bump'. The app detects potholes when people drive over them. This information has been used to allocate city resources in the fixing of potholes to areas with high prevalence of potholes. However Crawford (2013) points out that the distribution of signals from the app might be biased as many people from lower income neighbourhoods might not be able to afford a smartphone (note that this paper was written in 2013 and the app was used prior to this).

Kallus and Zhou (2018) address the problem of misrepresentation in the context of Police stop-and-frisk, where biased police behaviour leads to over-proportional stopping of racial minority group. They also show that adjusting a classifier for fairness while ignoring sample bias still leads to discriminatory classifications. That is, simple fairness adjustments to not account for biased data.

Finally, a dataset might induce bias in an algorithm if it is trained on **proxy variables**, i.e. features that are highly correlated with the protected group feature. We have discussed the practice of "blindness to sensitive attributes" in Section 2.3 and argued for the conscious accounting of sensitive attributes in the training data. Still, including these variables in the training of the classifier can lead to unfair outcomes. In this case we face the tradeoff that the criteria that are useful in making good classifications can also be very likely to induce discrimination across protected group features (Barocas and Selbst, 2016). This can be addressed, either by conducting a counterfactual analysis Kilbertus et al. (2017) and Kusner et al. (2017) (which would require a creative identification strategy) or by bayesian estimation techniques and the assumption that the prediction of the classifier is independent of the protected group feature if we condition on observed features (Jung et al., 2018a).

### 3.2  Bias in algorithm development

In addition to data bias as a source of unfairness, there are also biases introduced by wrong practices during the training of the algorithm. Throughout the development of a (fair) algorithm, researchers are faced with numerous decisions that could lead the outcomes to very different directions, e.g. the selection of the dataset, the selection and encoding of features selected from the dataset, the selection and encoding of the outcome variable, the rigour in identifying sources of bias in the data, the selection and specification of specific fairness criteria etc.. Every decision made contains an implicit assumption. Green (2018b) calls this "silent normative assumptions". They are silent because they seem hidden behind an adherence to mathiness and procedural rigour in the development of the algorithm (Green, 2018b). Moreover, in many cases the underlying assumptions can be normative. For instance, Dwork et al. (2012) pursue the fairness idea that similar people should be treated similarly but the function and feature that define similarity will have to be determined in the development of the fair classifier.

Barocas and Selbst (2016) shows how choosing and specifying the outcome variable relies on normative assumptions. This is also manifested in the specification of the "omitted payoff bias" in Kleinberg et al. (2017). The point here is that ADM systems for judges, like COMPAS, only evaluate the case along one dimension, namely re-offense risk. On the other hand, judges may aim to satisfy multiple goals in their ruling, such as preventing further crimes by the offender, deterring others from committing similar crimes, rehabilitating offenders, appropriate punishment, societal costs etc.. Thus, replacing a human decision maker by an ADM system could reduce the normative dimensions of rulings in criminal justice to a single goal causing a shift in the value grounds of the criminal justice system. This might go unnoticed if these 'silent' assumptions are not made open to relevant stakeholders for discussion (Green and Hu, 2018).

The overall procedure of developing algorithms that are adjusted to any one of the definitions of algorithmic fairness can be criticized as biased if it does not take into account the social and moral context of the decision made. For instance, the above mentioned Northpointe and ProPublica dispute due to the incompatibility of the respective fairness criteria (calibration and error balance) is rooted in the fact that black offenders have a higher re-offense risk than white offenders. Taking this fact as given when developing the algorithm would however ignore a long history of societal and institutionalized racism that has led to the present situation (Green and Hu, 2018).

Finally, we should take into account feedback and equilibrium effects. Feedback occurs when we change decision making processes and outcomes (e.g. through the implementation of ADM systems) which yields consequences that are not reflected in the data (of past decisions) that we use to train the ADM system on. Equilibrium effects are present when decisions on individuals also affect the composition of the group that the individual operates in (Barocas et al., 2018b). For instance, the decision of granting a loan has consequences on the person that is object to the decision and the people in its environment. This person can use this loan to open a successful business which will affect her credit score in the future. This might also affect the people in the environment of the business or the credit score of other individuals with the same characteristics (Liu et al., 2018). A jailed person will have a dent in its criminal history and consequently a harder time reintegrating into society after the jail sentence is served. A person that never gets a loan for a planned business might have a harder time improving its financial situation and will consequently most likely be denied credit in the future, too. Not taking these consequences and group effects into account will lead to a reinforcement of historical and group discrimination.

### 3.3 Representational harms

We have discussed sources of bias in the data as well as in the development of the algorithm in the previous sections. There are other forms of bias that are more difficult to sort into either one of these categories as they do not immediately lead to measurable unequal treatment or unequal outcomes for protected groups. This type of bias is more inherent in algorithms that affect our everyday lives, such as image search engines, translation tools or autocomplete algorithms for text messengers. For instance, Kay et al. (2015) provides evidence how image search results for occupation terms such as 'CEO' reflect (and even perpetuate) prevailing stereotypes and prejudices about the gender and race composition of such occupations. That is, search results for 'CEO' or 'Software developer' show mostly men. Another example is given by google translate. Translating the sentences "She is a doctor. He is a nurse." into a (in this case) gender-neutral language such as Turkish and then translating it back to English yields the result "He is a doctor. She is a nurse.". This mirroring of prejudices in algorithms that we use every day is termed 'representational' harms (Crawford, 2013). The problem with these harms is that these algorithms affect the environment we experience in our everyday lives causing us to assume that these stereotypical and prejudiced notions are the norm. The harms from this are more long-term and more subtle but they reinforce maltreatment and subordination of protected groups.

## 4   Challenges and research directions in algorithmic fairness

Research on fairness in machine learning still faces many challenges. In the formalization of fairness we face the problem that different definitions are useful in different contexts. These definitions cannot be unified in one generic framework as we face tradeoffs between individual definitions. We deal with these tradeoffs by embedding the fairness aspects of decisions that we want to address algorithmically in societal, institutional, legal and ethical frameworks on a case-by-case basis. We have to accept that there is no general once-size-fits-all solution to fairness in algorithms. Therefore, ADM system developers will have to **collaborate with domain experts**. For this purpose, ADM systems also have to be **explainable**.

There are two promising ways to avoid unfairness due to biased data. One is to fully **understand the data used**. The other is to **improve the quality of the data** used. Methods of **causal inference** will become more relevant in order to understand the underlying mechanisms of a decision process as well as identify sources of bias in data (Kleinberg et al., 2017; Jung et al., 2018a). This goes in line with improving the knowledge of researchers on the data used as well as increasing awareness for the importance of understanding the data-generating process (Gebru et al., 2018). In addition, a **focus on data diversity** is crucial to address problems of discrimination against minorities (Drosou et al., 2017).

Implementing dynamic feedback and equilibrium effects into models of decision making is necessary to understand the consequences of changing decision processes (Liu et al., 2018; Corbett-Davies and Goel, 2018). Further research on real world applications of ADM systems is necessary to **understand how human decision makers interact with machine support** (Green, 2018a). This also means that algorithms will have to undergo **frequent reevaluations**. No algorithm can be fair forever without readjustments. Again, increased diversity becomes crucial when it comes to subtle representational harms. In this case, not only the diversity in the data is important but also **diversity in the group of ADM system developers** (Crawford, 2017).

## 5   Conclusion

This report reviews the literature on fairness in machine learning. Fairness in algorithmic decision making is crucial as algorithms affect more and more sensitive areas where decisions affect individual lives and society in general. We discuss formal specifications of algorithmic fairness, their strength and weaknesses as well as the tradeoffs among them. Ensuring fairness in machine learning is no trivial task. This is because we are trying to implement a complex social construct into a metric that is understandable to machines. Ignoring the complexities of fairness that cannot be expressed in metrics is doomed to end up in unfairness. Finally, we discuss future research directions that seem promising in addressing the problems that arise in the development of fair algorithms. It is apparent that there will never be a simple one-size-fits-all solution to this. Research must increasingly and explicitly take into account the social and institutional context and keep track of the consequences that arise from the implementation of the algorithm. (Gomez, 2018)

This is costly but there is also a merit in this as it ensures that decisions become accountable and it forces the discussion on fairness in decision making to become transparent. Thus, transparency, which is not mere access to source code but a degree of algorithmic explainability that enables humans to understand and challenge algorithmic decisions"(Gomez, 2018) is crucial to achieve fairness in algorithmic decision making.